\documentclass[sigconf,authorversion]{acmart}
\AtBeginDocument{%
  \providecommand\BibTeX{{%
    \normalfont B\kern-0.5em{\scshape i\kern-0.25em b}\kern-0.8em\TeX}}}

\copyrightyear{2020}
\acmYear{2020}
\setcopyright{acmlicensed}\acmConference[ETRA '20 Short Papers]{Symposium on Eye Tracking Research and Applications}{June 2--5, 2020}{Stuttgart, Germany}
\acmBooktitle{Symposium on Eye Tracking Research and Applications (ETRA '20 Short Papers), June 2--5, 2020, Stuttgart, Germany}
\acmPrice{15.00}
\acmDOI{10.1145/3379156.3391340}
\acmISBN{978-1-4503-7134-6/20/06}

\begin{document}

\title{MutualEyeContact: A conversation analysis tool with focus on eye contact}


\author{Alexander Schäfer}
\affiliation{%
  \institution{TU Kaiserslautern}
  }
\email{alexander.schaefer@dfki.de}

\author{Tomoko Isomura}
\affiliation{%
  \institution{Waseda University \\ Japan Society for the Promotion of Science}
}
\email{isomurat8818@gmail.com}

\author{Gerd Reis}
\affiliation{%
  \institution{German Research Center for Artificial Intelligence}
  }
\email{gerd.reis@dfki.de}

\author{Katsumi Watanabe}
\affiliation{%
  \institution{Waseda University \\University of New South Wales}
}
\email{katz@waseda.jp}

\author{Didier Stricker}
\affiliation{%
  \institution{German Research Center for Artificial Intelligence \\TU Kaiserslautern}
  }
\email{didier.stricker@dfki.de}

\renewcommand{\shortauthors}{Schäfer et al.}

\begin{abstract}
Eye contact between individuals is particularly important for understanding human behaviour. To further investigate the importance of eye contact in social interactions, portable eye tracking technology seems to be a natural choice. However, the analysis of available data can become quite complex. Scientists need data that is calculated quickly and accurately. Additionally, the relevant data must be automatically separated to save time. In this work, we propose a tool called MutualEyeContact which excels in those tasks and can help scientists to understand the importance of (mutual) eye contact in social interactions. We combine state-of-the-art eye tracking with face recognition based on machine learning and provide a tool for analysis and visualization of social interaction sessions. This work is a joint collaboration of computer scientists and cognitive scientists. It combines the fields of social and behavioural science with computer vision and deep learning.
\end{abstract}





\maketitle

\section{Introduction}
\begin{figure}
  \centering
  \includegraphics[width=1\linewidth]{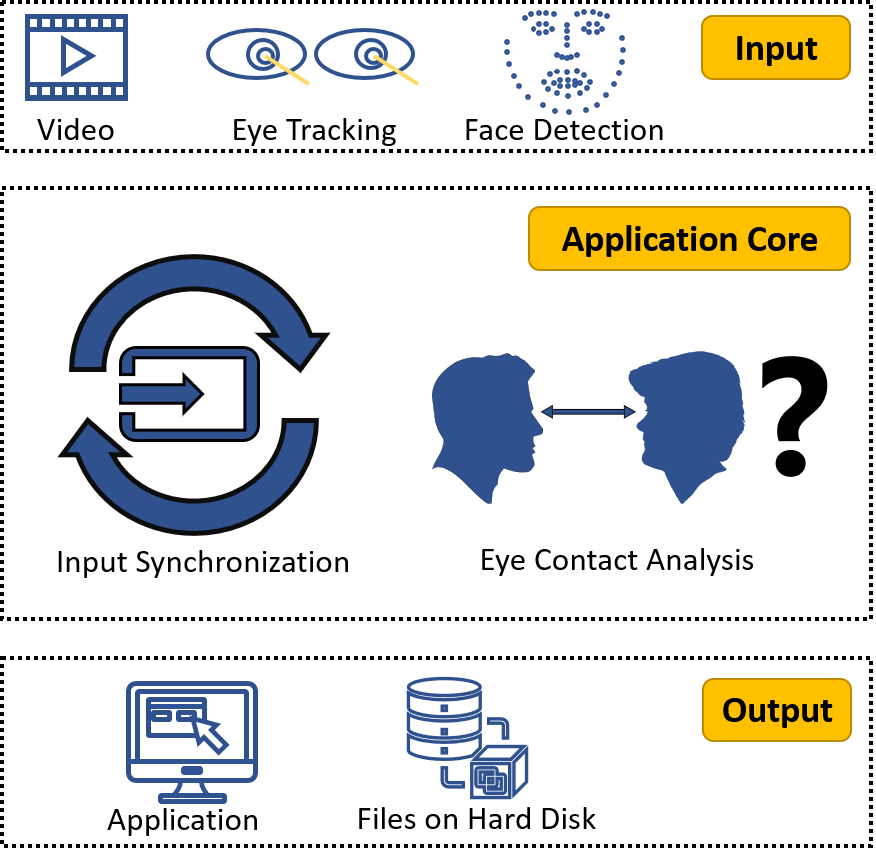}
  \caption{MutualEyeContact is a tool that combines state-of-the-art tracking and detection algorithms. Main focus is the analysis of conversations between two individuals in respect to eye contact.}
  \Description{Graphical User Interface of the proposed tool.}
  \label{fig:main}
\end{figure}

Scientists need to be able to analyse collected data efficiently and correctly in order to draw useful conclusions. In our case we are addressing the problem of efficiently analysing human behavior and emotions during social interactions between individuals.
We want to analyse how the human body behaves during eye contact with another person and to the best of our knowledge, there is no specific tool available to analyse this case efficiently. MutualEyeContact was developed for this purpose (see Figure \ref{fig:main} for an overview). It supports scientists in understanding the behavior of the human body during natural social interactions. We combine and synchronize machine learning based face tracking with wearable eye tracking hardware. 

Researchers have spent many hours analysing and labeling videos by hand because there is no tool available for their specific research problem. As an example, \cite{rogers2018using} reported that manual coding of gaze behavior in 4-min conversations took them approximately 62 hours. It is worth mentioning that subjects need some time to get used to invasive tracking devices. This can lead to falsified data in the beginning of a recording session since participants are behaving unnatural due to the observation. To compensate for this falsified data, longer data recording sessions should be preferred. This increases the necessary manual labeling work even more. Therefore, short sessions for manual labeling tasks are currently preferred. By utilizing the technique of a vision-based face-tracking system, most of the time spent for this process can be omitted. Our tool is automated and therefore does not add any additional work depending on the amount of available data. 

For eye tracking we are using Tobii Pro Glasses 2 (TPG2). OpenFace 2.0 toolkit from \cite{baltrusaitis2018openface} is used for facial landmarks and facial action unit recognition. Facial landmarks represent important regions of the face such as eyes, eyebrows, nose, mouth and jaw. The Facial Action Coding System (FACS) \cite{friesen1978facial} is a coded system to support emotion analysis. It assigns Action Units to almost every visible movement of the musculature used for face mimic.

Our work fosters in-depth analysis of data related to eye contact in natural social interactions. The contribution of this work can be summarized as follows:

\begin{itemize}
    \item Significant time savings by using the presented tool for mutual eye contact analysis in respect to traditional methods
    \item More in-depth analytics for researchers due to automatic data extraction while analysing recorded footage with filter selection (e.g. mutual eye contact + Facial Action Units)
\end{itemize}

\section{Background and Related Work}
\label{chap:background}
Eye-gaze plays various roles in human social interactions. In particular, through the direct gaze which often results in eye-contact, we are able to perceive and signal a variety of meanings, such as intention to communicate or to exchange turns of speaker \cite{ho2015speaking}, threat and dominance \cite{emery2000eyes}, interest \cite{argyle1976gaze}, or seeking for approval \cite{efran1968looking}. Recently, some theories propose that eye-contact activates social brain networks \cite{senju2009eye}, and facilitates self-referential processing \cite{conty2016watching}. Furthermore, the latest hyper-scanning studies showed that eye-contact triggers neural coherence between agents \cite{hirsch2017frontal}, \cite{piazza2018infant}, suggesting that eye-contact enhances the temporal alignment of two brains and facilitates the information sharing \cite{canigueral2019role}. While socio-cognitive function of eye-contact has frequently been shown, many of these studies are conducted under experimentally manipulated settings, where for example the participants are required to intentionally fixate on the partner’s eye region. One of the reasons for this is because tracking eye-gaze in natural human-to-human interactions is by no means easy. In real human-to-human interactions, such as dyadic conversation, eye movements of both agents are generally very active. Since both agents constantly alternate their gaze between eyes of their partner and other regions, the exchange of the eye-gaze happens very quickly. Accordingly, temporal and spatial resolution of measures is critical to address the interpersonal dynamics of gaze behavior in naturalistic settings.

Recently, several studies have tackled the nature of temporal dynamics of gaze in real human-to-human interactions by simultaneously measuring both agent’s eye-gaze using wearable eye-trackers \cite{ho2015speaking, rogers2018using, rogers2019contact, broz2012mutual}. In particular, Rogers et al. (2018) used eye tracking and attempted to code the temporal and spatial details of each agent’s gaze patterns and its interpersonal interactions during a 4-min natural conversation, by analysing the time course of fixated facial area (e.g., forehead, eyes, nose, mouth). They showed that during the conversation the agents spent on average 60\% of the time directing their gaze toward face of the other person, but only 10\% of time directed specifically to the eyes; and that mutual face-gaze events were approximately 2.2 s long, but it was only 0.36 s long for the mutual eye-contact events. These results demonstrated that occurrence of mutual eye-contact is surprisingly momentary.

While the techniques of dual eye-tracking seemingly paves the way for cognitive scientists to further investigate the role of eye-contact in natural social interactions, load for manually coding the gaze allocation apparently prevents the efficient progress of the research. Wearable eye-tracking glasses usually have two cameras; one facing forward to record a video of wearer’s field of vision; and one measures the wearer’s eye gaze which is represented as a 2-D coordinate based on a frame-by-frame static image of the video. As the faces of both agents constantly move during a natural conversation, coordination of the face and features within the face do not remain in a fixed location in the video, meaning that gaze allocation should be identified frame-by-frame based on both video and gaze data. As dual eye-tracking requires twice of this work, a substantial amount of time is needed to manually code the gaze behavior of both agents. 

\section{Implementation}
In this section we describe how MutualEyeContact is implemented. We explain how multiple systems are integrated and how the synchronization of video, eye tracker and face detection was done. The application allows to apply filters (e.g. mutual eye contact, eye contact) to automatically extract and display useful information from the input. Action Units (blinking, chin raiser, brow raiser etc.) provided by the face recognition algorithms can be extracted frame-by-frame as well (for a full list of available Action Units we refer the reader to \cite{baltrusaitis2018openface}, \cite{baltruvsaitis2015cross}).

Filters can be combined with each other to even further highlight certain aspects of input data. For example, it is possible to combine Action Unit filters and estimate emotions, e.g. \textit{Cheeck Raiser} + \textit{Lip Corner Puller} is considered as happiness emotion according to EMFACS (Emotional Facial Action Coding System \cite{friesen1983emfacs}). This filter could then be used find a correlation between eye contact and the happiness emotion.

\subsection{Data Filtering}
\begin{figure}[h]
  \centering
  \includegraphics[width=\linewidth]{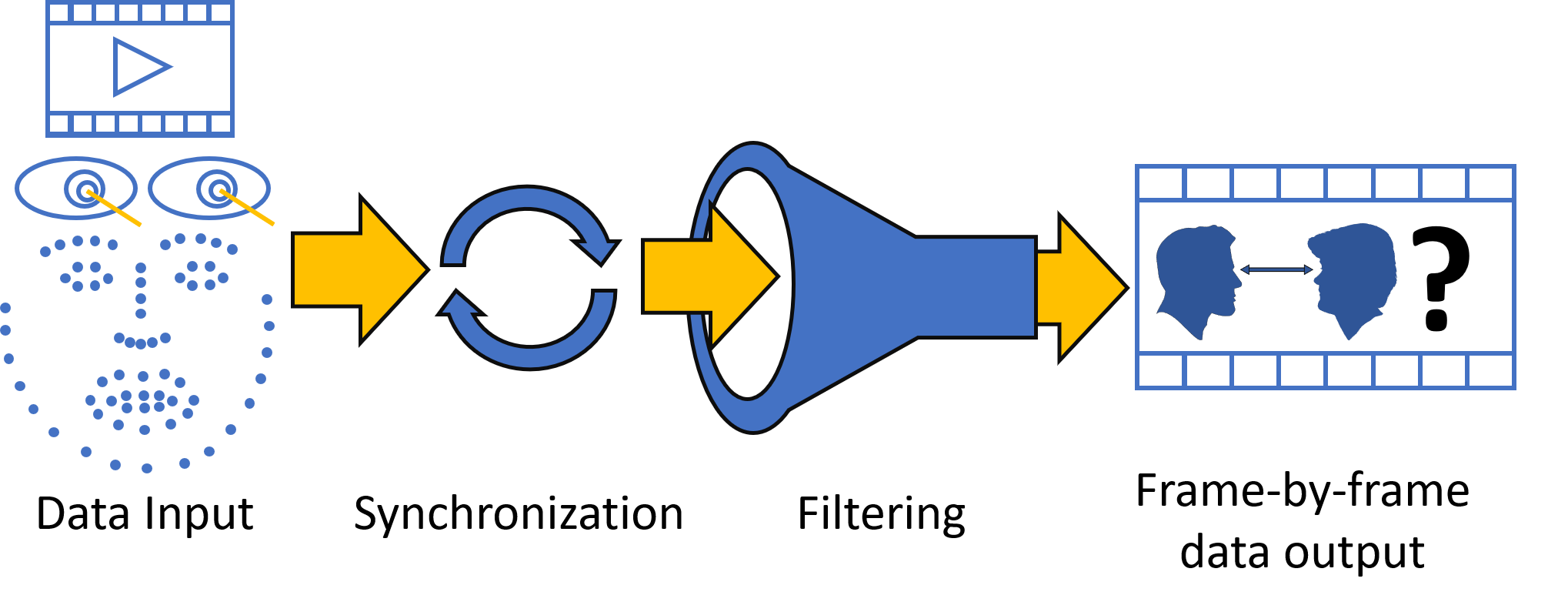}
  \caption{Schematic overview of our data processing pipeline. The tool takes video, eye tracking and face detection data as input. After the data is synchronized, filters can be applied to the data to highlight certain video parts.}
  \Description{Schematic overview of our data processing pipeline. The tool takes video, eye tracking and face detection data as input. After the data is synchronized, filters are applied to the data to highlight certain video parts. For example, video frames where mutual eye contact between two individuals is established.}
  \label{fig:filter1}
\end{figure}

Most filters will process a whole video, extracting eye tracking and face recognition data at a given frame and calculate specific data with a specified output as shown in Figure \ref{fig:filter1}. This will produce a frame-by-frame data output which can be used to analyse behaviour at certain areas of the input data. Some filters are used to combine others e.g. mutual eye contact filter uses two eye contact filters.

As an example, there is a recorded conversation between person A and person B. Both will have their eye gaze tracked by wearable eye tracking devices. Additionally to that, the outward facing camera of the eye tracker is recording the field of view of person A, which is used for face detection of person B and vice versa. If a scientist wants to know whether person A is looking into the eyes of person B, the eye contact filter can be applied to the input data. This filter will extract eye gaze position $\mathit{A_{eye}}$ of person A and face position $\mathit{B_{face}}$ of person B for each video frame by using the internal synchronization algorithms (see chapter \ref{chap:sync}). 
After $\mathit{A_{eye}}$ is obtained, the pixel coordinate $\mathit{P(x,y)}$ of the eye gaze is obtained by re-projecting $\mathit{A_{eye}}$ to the image plane of $\mathit{B_{face}}$. After that a point in polygon test with point $\mathit{P(x,y)}$ and the landmarks of $\mathit{B_{face}}$ is done. This results in a \em true \em or \em false \em value for each frame, which is stored in an internal data structure. If required, a more descriptive and continous value in the range of 0 to 1 of eye/face contact can be calculated by using the known distance between gaze point and the region of interest (i.e. eye or face area). 

After a filter is completed and has finished processing, it is displayed in a highly customizable timeline widget (see Figure \ref{fig:filter}).

\subsection{Input Data Synchronization}
\label{chap:sync}
As mentioned in chapter \ref{chap:background}, face-gaze events are approximately 2.2s long, while mutual eye-contact lasts for 0.36s, represented by 9 consecutive frames in the video. Therefore, the synchronization deviation should not exceed 9 frames in case of a 25Hz video. With the approach described in this section, we can guarantee a synchronization accuracy of less than 3 frames. Eye trackers usually have a higher framerate than videos, in our case we used trackers with 50Hz. The eye tracker uses inward facing cameras for eye tracking and one additonal outward facing camera to show what the wearer is seeing. The eye tracking data is then re-projected to an image plane of the outward facing camera where the eye gaze is available as pixel coordinates. The data stream from the eye tracker needs to be synchronized to the outward pointing camera and any additional data (e.g. face recognition). The face recognition output from OpenFace toolkit uses OpenCV \cite{opencv_library} in the backend and produces an output which labels each individual frame with a unique number. The eye tracking data stream consists of a timestamp $\mathit{T_{e}}$ from the eye Tracker and a corresponding video time stamp $\mathit{T_{vts}}$ which represents the current time on the tracker since the video started.

Since frame encoding/decoding does vary depending on the used codec and OpenCV not using video timestamps as frame descriptor, there is a descriptor mismatch problem which needs to be solved. Additionally, there is also a framerate mismatch, because the eye tracker is having a much higher framerate ($50Hz$) than the video ($25Hz$). 


The eye tracker sends a periodic signal which contains a timestamp from the eye tracker $\mathit{T_{e}}$, a corresponding video time stamp $\mathit{T_{vts}}$ and a presentation timestamp $\mathit{T_{pts}}$. Presentation timestamps (PTS) are used in video synchronization and denote a start time $\mathit{PTSB}$ and an end time $\mathit{PTSE}$ when an individual frame should be displayed. With this information, we can solve the mismatch problems with equation \ref{eq:one}.
\begin{equation}
    F_{number} = \frac{F_{ptse} - FF_{ptse}}{F_{dur}}
    \label{eq:one}
\end{equation}
Here $F_{ptse}$ denotes the PTS end time of the current frame, $FF_{ptse}$ the PTS end time of the first valid frame in the video and $F_{dur}$ denotes the time in microseconds a frame takes to be displayed.

We used the uncompressed video output from the eye tracker as input to our software to avoid unnecessary information loss due to encoding/decoding. For displaying the video frames we chose the Windows Media Foundation (WMF) library for video decoding since it provides accurate video timestamps for each frame. 

\subsection{Tool Overview}
\begin{figure*}[h]
  \centering
  \includegraphics[width=\linewidth]{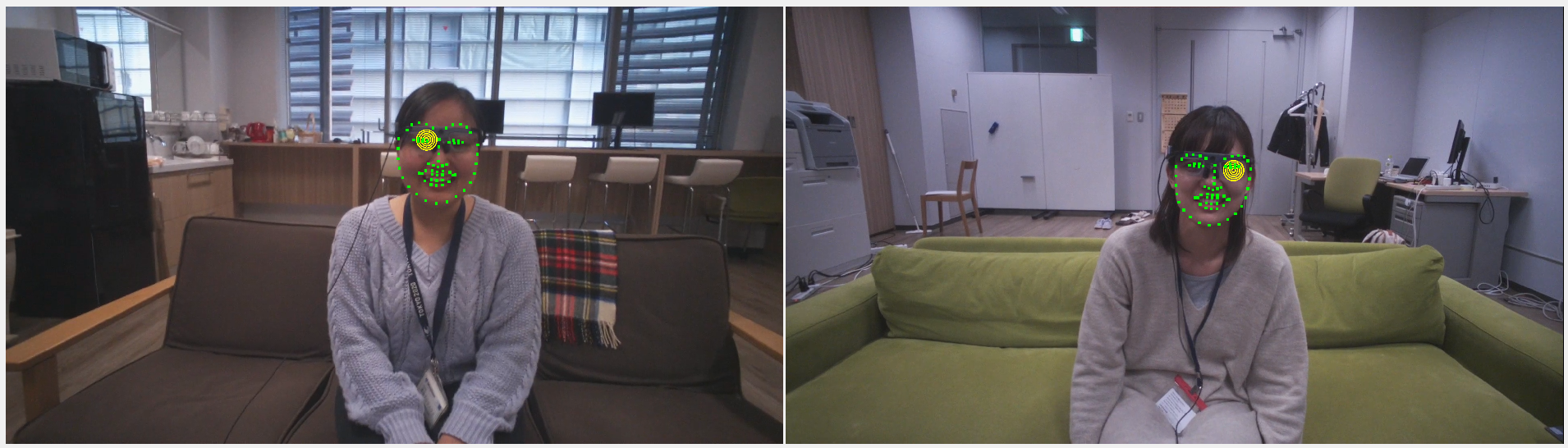}
  \caption{Side-by-side view of two input videos as shown in the proposed tool. The green dots are highlighting the facial landmarks detected by the face detector. The yellow circle represents the gaze point as tracked by the eye tracker}
  \Description{Graphical User Interface of our proposed tool. The top shows two input videos with controls while the bottom shows filters which are applied to the input data.}
  \label{fig:gui1}
\end{figure*}

In this section we briefly introduce the graphical user interface (GUI) of the proposed tool as well as some suggestions which were implemented after testing the tool. The GUI consists of four main elements: data loading, video viewing, video controls and filters. The video viewing part of the video features a side-by-side view of two input videos as shown in Figure \ref{fig:gui1}. Green dots in the area surrounding the face are facial landmarks detected by the face detector. A yellow circle represents the gaze point of the agent wearing the eye tracker.

\begin{figure}[h]
  \centering
  \includegraphics[width=\linewidth]{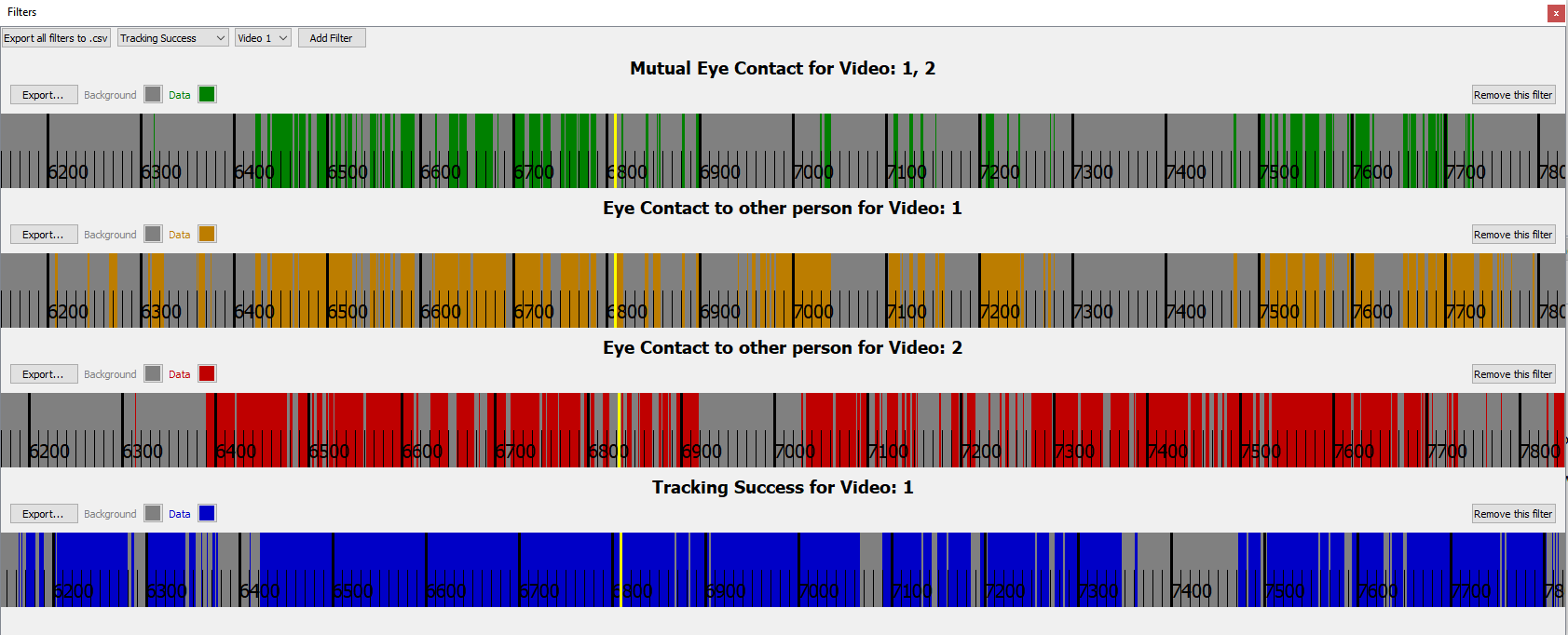}
  \caption{Multiple filters applied to the input data, each represented in it's own timeline widget.}
  \Description{Highlighting a specific area in the video for mutual eye contact data.}
  \label{fig:filter}
\end{figure}

After loading the necessary files and pre-processing (e.g. solving the synchronization problem), the controls for video manipulation are enabled. As described in the previous section, filters can be applied to the input data which is the core aspect of our tool. There can be multiple filters applied to the data, each represented by a new timeline widget as shown in Figure \ref{fig:filter}. Applying filters to the data consumes a lot of processing power, depending on how much data is available (e.g. a 20 minute video has approx. 30.000 frames at 25Hz). Processing time can vary from a few seconds to a few minutes depending on the length of a recorded session. For a video with about 30.000 frames, the eye contact filter requires about 0.5 seconds and the mutual eye contact filter about 1.5 seconds to compute (on an Intel Xeon E3-1245 v6 CPU). For this reason, each time a filter is applied, a new thread is created that processes the data simultaneously. While a filter is being processed, a content blocker is shown over the specific filter area. This enables a smooth non-blocking user experience for the user. Hardware acceleration for video display is used, since multiple videos in at least $1920x1080$ resolution can result in a staggering view experience.

\section{Pilot Testing and Experiments}
To evaluate our tool, computer scientists as well as cognitive scientists tested the application. We previously recorded multiple social interaction sessions as data input. To record this data, participants were told to sit in front of each other with a distance of about 3 meters. Each participant was wearing an eye tracker to record the eye gaze. The outward facing camera of the eye tracker recorded the field of view of the participant. The participants were told to have a conversation for about 20 minutes. After that the session closed and the data was stored.

\begin{figure}[h]
  \centering
  \includegraphics[width=\linewidth]{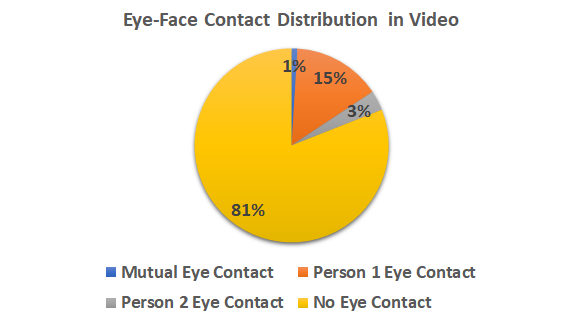}
  \caption{Eye-Face contact distribution during a recorded 20 minute social interaction session.}
  \Description{Eye contact distribution during a recorded 20 minute social interaction session. This data can be easily extracted with the proposed tool.}
  \label{fig:diagram}
\end{figure}

We used our tool to explore certain data during the social interaction sessions. As an example, we calculated the eye-face contact distribution which is shown in Figure \ref{fig:diagram}. This data is a combination of several filters i.e. \textit{mutual eye contact}, \textit{eye contact person 1} and \textit{eye contact person 2}. It should be mentioned that face recognition does not work in all circumstances (e.g. due to occlusion) and frames without valid face recognition are therefore omitted. While a manual dissection of this input data would take many hours, our tool requires only 1-3 seconds to compute.
\section{Conclusion and Future Work}
In this paper, we proposed a tool for researchers to analyse the various roles of eye contact in natural social interactions. By combining reliable eye tracking with face detection we are able to create a useful tool for scientists that can save many hours of labeling and analysing videos manually. Wearable eye tracking devices are used for reliable and accurate eye-gaze tracking. We use facial landmark and facial Action Unit recognition together with eye-gaze tracking and combine it into one system. This tool fills the gap between the fields of social and behavioural science, computer vision and machine learning to enable a powerful in-depth analysis for eye contact related research problems. In the future, this work can be extended with features like remote photoplethysmography for non-invasive vital sign monitoring during social interactions (e.g. heart rate, respiration rate). We also plan to include monocular full body 3D pose estimation as described in the work of \cite{kovalenko2019structure} to even further support the investigation of the human body during social interactions. In future work, we intend to use this tool to conduct (mutual) eye contact research studies.

\begin{acks}
This work was partially funded by: Offene Digitalisierungsallianz Pfalz which is part of the Innovative Hochschule (Grant No: 03IHS075B); CREST, Japan Science and Technology Agency (Grant No: PMJCR14E4); KAKENHI, Japan Society for the Promotion of Science (Grant No: 17H06344 and 19K20651); VIDETE, German Federal Ministry of Education and Research (BMBF) (Grant No. 01|W18002).
\end{acks}

\bibliographystyle{ACM-Reference-Format}
\bibliography{sample-base}


\end{document}